\documentclass[journal, twocolumn, web]{article}
\usepackage{tmi}
\usepackage{cite}
\usepackage{amsmath,amssymb,amsfonts}
\usepackage{algorithmic}
\usepackage{graphicx}
\usepackage{textcomp}

\usepackage{xspace}
\usepackage{xcolor}
\usepackage{multirow}
\usepackage{adjustbox}
\usepackage{svg}
\usepackage{booktabs}
\usepackage{array}

\newcolumntype{L}[1]{>{\raggedright\arraybackslash}p{#1}}

\newcommand{\ceil}[1]{\lceil #1 \rceil}
\def\BibTeX{{\rm B\kern-.05em{\sc i\kern-.025em b}\kern-.08em
    T\kern-.1667em\lower.7ex\hbox{E}\kern-.125emX}}
\begin{document}
\title{Robust Surgical Phase Recognition From Annotation Efficient Supervision}
\author{Or Rubin, and Shlomi Laufer
\thanks{This study was funded by The Bernard M. Gordon Center for Systems Engineering at the Technion.}
\thanks{
The authors are with the department of Data and Decision Sciences, Technion, Israel Institute of Technology, Haifa 3200, Israel. (e-mail: orrubin@campus.technion.ac.il;$\quad$ laufer@technion.ac.il). 
    }
}
\date{}

\maketitle

\label{sec:abstact}
\begin{abstract}
Surgical phase recognition is a key task in computer-assisted surgery, aiming to automatically identify and categorize the different phases within a surgical procedure. Despite substantial advancements, most current approaches rely on fully supervised training, requiring expensive and time-consuming frame-level annotations. Timestamp supervision has recently emerged as a promising alternative, significantly reducing annotation costs while maintaining competitive performance.  However, models trained on timestamp annotations can be negatively impacted by missing phase annotations, leading to a potential drawback in real-world scenarios.  In this work, we address this issue by proposing a robust method for surgical phase recognition that can handle missing phase annotations effectively. Furthermore, we introduce the SkipTag@K annotation approach to the surgical domain, enabling a flexible balance between annotation effort and model performance. Our method achieves competitive results on two challenging datasets, demonstrating its efficacy in handling missing phase annotations and its potential for reducing annotation costs. Specifically, we achieve an accuracy of 85.1\% on the MultiBypass140 dataset using only 3 annotated frames per video, showcasing the effectiveness of our method and the potential of the SkipTag@K setup. We perform extensive experiments to validate the robustness of our method and provide valuable insights to guide future research in surgical phase recognition. Our work contributes to the advancement of surgical workflow recognition and paves the way for more efficient and reliable surgical phase recognition systems.
\end{abstract}


\section{Introduction}
\label{sec:introduction}
Surgical data science \cite{maier2017surgical, maier2022surgical} has great importance in the field of computer aided surgery. Surgical phase recognition is an important task with a growing interest in recent years \cite{demir2023deep}, focusing on automatically identifying and categorizing the different phases within a surgical procedure. This task has various applications, such as automatic indexing of surgical video databases \cite{twinanda2016endonet} and skill assessment of surgeons \cite{Liu_2021_CVPR, komatsu2024automatic}.
\\
Despite the substantial advancements in surgical phase recognition \cite{demir2023deep}, most current approaches rely on fully supervised training, requiring frame-level annotations for all videos in the training set. Obtaining such a training set is both time-consuming and costly, as it necessitates experienced surgeons to meticulously annotate the videos. This issue may hinder the adoption of automatic surgical phase recognition tools for new surgical procedures and use cases. To address this challenge, Ding \etal \cite{ding2023less} proposed a promising direction based on timestamp annotations, where only a single frame per phase is annotated. This approach has been shown empirically to reduce annotation time by 74 \% compared to full annotation while still achieving comparable results.

However, when dealing with complex surgeries where the order of phases is not deterministic, a potential drawback of timestamp annotation is the possibility of missing phases that were overlooked by the annotator. In this work, we delve into this issue and explore the robustness of the model in relation to such missing phases, proposing a method that is robust to missing phase annotations.

Furthermore, we introduce the SkipTag@K annotation method to the surgical domain, where only K frames from a video are annotated. This method has immense potential in terms of annotation time efficiency. By employing SkipTag@K, we achieve an accuracy of 83.6 on the Cholec80 dataset \cite{twinanda2016endonet} using only 2 samples per video, significantly reducing the annotation burden. 

In summary, our main contributions are as follows:
\begin{enumerate}
    \item We address the problem of missing labels in timestamp supervision and present a robust method to handle this issue.
    \item We introduce SkipTag@K to the surgical domain and demonstrate its potential for efficient annotation.
    \item We provide in-depth empirical studies of our method across two challenging datasets in the surgical domain.
\end{enumerate}

Our code will be made available 
upon acceptance.


\section{Related Work}
\label{sec:related_work}
\subsection{Surgical phase recognition}
Surgical phase recognition has been explored using various approaches across several types of surgeries, including cataract surgeries \cite{zisimopoulos2018deepphase}, laparoscopic cholecystectomies and \cite{twinanda2016endonet} TKR procedures \cite{kadkhodamohammadi2021towards}. Early works have relied on hand-crafted features and statistical models \cite{blum2010modeling}, \cite{padoy2012statistical}. Recent works leverage the capabilities of deep learning for surgical phase recognition, 
often adopting a two-stage network architecture. 
These architectures initially extract features using backbone networks such as ResNet \cite{he2016deep},
Inception \cite{szegedy2016rethinking}, or Vision Transformers (ViT) \cite{dosovitskiy2020image}. Subsequently, temporal dependencies are modeled using an additional model such as Long Short-Term Memory (LSTM) \cite{jin2017sv}, 
Multi-Stage Temporal Convolutional Networks (MS-TCN) \cite{czempiel2020tecno}, and Transformer networks \cite{czempiel2021opera}. 
These two-stage approaches aim to effectively capture spatial and temporal information from the surgical videos. 
While most works focus on the time-consuming fully supervised setup \cite{demir2023deep}, Several works explored several alternatives. In the semi-supervised setup, a partial subset of videos are fully annotated, and the rest remain unannotated. In this direction, Ramesh et al. \cite{ramesh2023dissecting} trained self-supervised feature extractors using four different methods, including the DINO method and evaluated the models in both fully-supervised and semi-supervised setups. They trained their models exclusively on the Cholec80 dataset and demonstrated the results on Cholec80, as well as the models' generalization abilities on other datasets. In our work, we adapt this training methodology to each dataset separately, namely the Cholec80 and MultiBypass140 datasets.
Additional works also perused the semi-supervised direction \cite{shi2021semi,yu2018learning,chen2018semi,yengera2018less}.
Other explored directions in the surgical domain include active learning \cite{shi2020lrtd}, \cite{bodenstedt2019active}, and timestamp supervision \cite{ding2023less}.

While active learning methods select clips or entire videos for annotation, our work focuses on setups where single frames are annotated, such as timestamp supervision and SkipTag@k. Single-frame annotations are less time-intensive compared to annotating long clips or entire videos, making them more efficient for obtaining labeled data. In contrast to active learning, where the model iteratively selects samples for annotation, SkipTag@k allows the selection of all samples at the same time, further simplifying the annotation process.
Several works have previously explored the use of pseudo-lables \cite{zhang2022retrieval, ward2021automated}. 
\cite{yu2018learning}. They first trained an offline model on a limited supervised training set, in order to generate pseudo-labels on the entire training set, then trained an online model on those pseudo-labels. 
\subsection{Action segmentation}
Surgical phase recognition can be viewed as a specialized case of the action segmentation task. 
The key differences lie in the unique characteristics of the surgical domain.
Several weakly supervised setups have been explored in relation to the action segmentation task, 
such as timestamp supervision \cite{li2021temporal,behrmann2022unified,khan2022timestamp, ijcai2023p77}, 
set supervision \cite{li2021anchor}, 
and transcript-based supervision where Huang \etal \cite{huang2016connectionist} suggested an extension to the Connectionist Temporal Classification (CTC) loss that utilizes the similarity between consecutive frames.
The challenge of missing actions has also been highlighted by prior works, such as Souri \etal \cite{Souri_2022_BMVC} which suggests an optimization-based pseudo-label expansion mechanism, and Rahaman \etal \cite{rahaman2022generalized} which proposes an EM-based approach. 
Rahaman \etal also introduce the SkipTag setting, where a fixed number of frames (K) are annotated per video. However, they only evaluate their method with K set to the average number of actions per video.
We extend this concept to SkipTag@K and evaluate our method using different K values.
Unlike Rahaman \etal's method, our approach does not assume a prior on the action lengths.
The semi-supervised setup was also explored in the action segmentation domain \cite{singhania2022iterative}.
\subsection{Imbalanced data}

The distribution of surgical phases is often highly imbalanced, with some phases occurring more frequently or lasting longer than others. This imbalance can lead to biased models that struggle to accurately recognize less prevalent phases. Various techniques have been proposed to address the class imbalance problem. Focal loss \cite{lin2017focal}, originally introduced for object detection, has been widely adopted in many domains, including the surgical domain. Zhang \etal \cite{zhang2021surgical} employed an unweighted focal loss for surgical phase recognition, while Ramesh \etal \cite{ramesh2021multi} utilized a class-weighted loss to mitigate the impact of imbalanced data. 
Weighted focal loss has also been applied in the surgical domain, albeit for surgical image classification tasks rather than phase recognition \cite{le2020transfer, qin2018weighted}.

\subsection{Uncertainty Estimation}
Monte Carlo Dropout (MCD) \cite{gal2016dropout} is a widely used method for estimating uncertainty in deep learning models by performing multiple forward passes with dropout enabled during inference, and it has a solid theoretical basis. Temperature scaling \cite{guo2017calibration} is another approach that calibrates the confidence scores of a trained model by introducing a temperature parameter to the softmax function.
In surgical phase recognition, Bodenstedt \etal \cite{bodenstedt2019active} employed MCD with several estimators, including an entropy-based estimator for uncertainty estimation in an active learning framework. Ding \etal \cite{ding2023less} leveraged MCD for uncertainty estimation using the standard deviation of predictions as a measure of uncertainty to generate reliable pseudo-labels for training.

\section{Methods}
\label{sec:methods}

\begin{figure*}
  \centering
  \includegraphics[width=\textwidth]{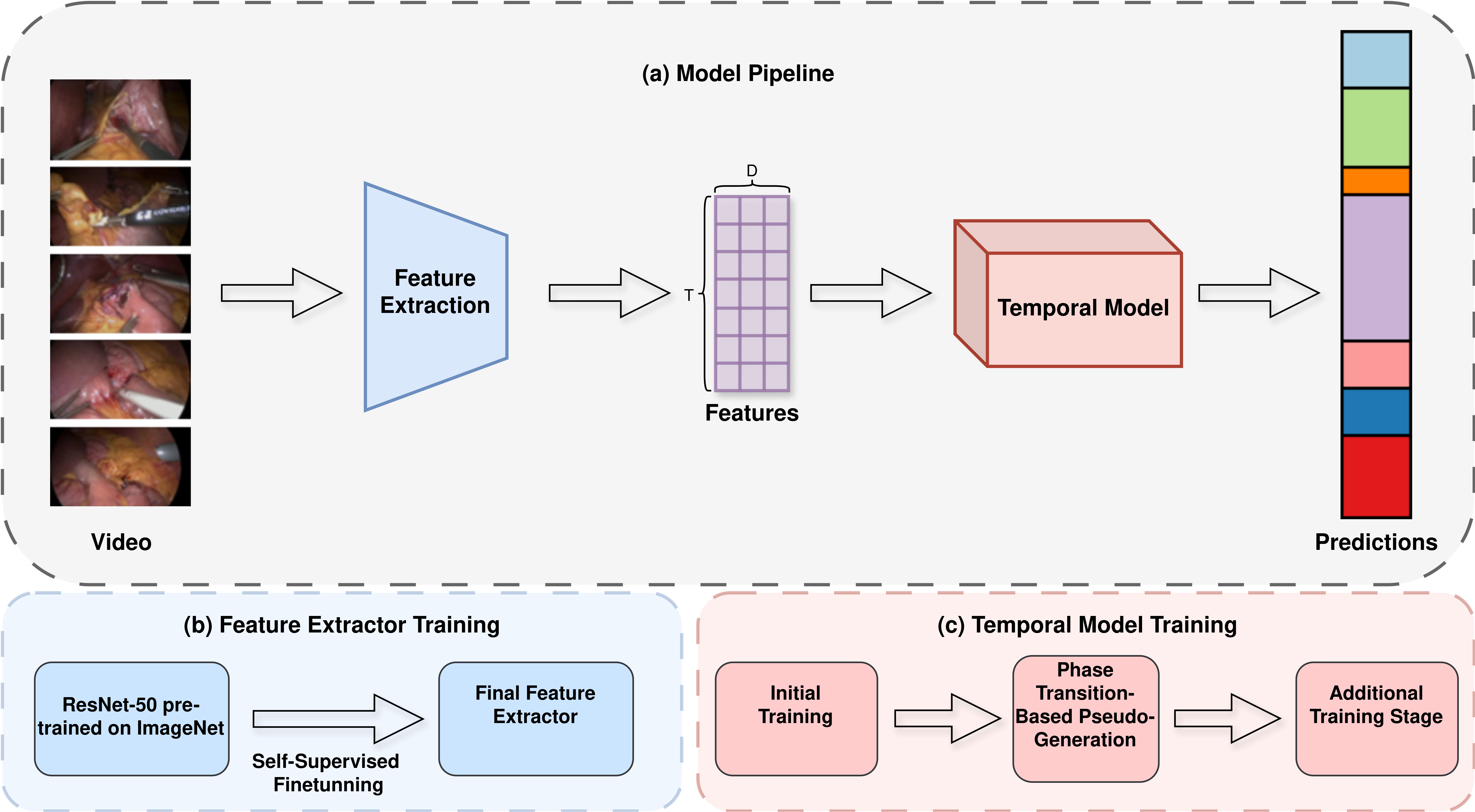}
    \caption{Overview of our proposed surgical phase recognition method. (a) The model prediction pipeline for generating phase predictions from an input surgical video. It consists of a two-stage architecture: a feature extraction model followed by a temporal model. (b) The feature extractor training pipeline. A ResNet-50 model pre-trained on ImageNet is fine-tuned using self-supervised learning on the target surgical video dataset. (c) The temporal model training pipeline. It includes an initial training stage using our proposed loss function to create a base model. The base model then generates pseudo-labels which are used to train the final temporal model.}
  \label{fig:Scheme}
\end{figure*}

The prediction and training schemes are illustrated in Fig.\ref{fig:Scheme}.

\subsection{Problem Formulation}
Let $X = [x_1, \ldots, x_T]$ be a surgical video containing $T$ frames. Our goal is to predict $\hat{Y} = [\hat{y_1}, \ldots, \hat{y_T}]$, which contains a phase classification for each frame in the video from a set of phase options $\mathcal{P} = \{1, \ldots, C \}$, where $C$ is the number of phase options. We have a set of VN videos $\overline{X} = \{X_i\}_{i=1}^{\text{VN}}$ with corresponding annotations.
The annotation type varies depending on the supervision setup. In the fully-supervised setup, every video $X \in \overline{X}$ of length $T$ has a corresponding labeling $Y_{full}=[y_1, \ldots, y_T]$. In the timestamp supervision setup, only one frame from each phase is annotated, resulting in a labeling $Y_{ts} = [y_{t_1}, \ldots, y_{t_{\text{PN}}}]$, where PN is the number of phases. In the SkipTag@$K$ supervision, a subset of $K$ frames is chosen and annotated, resulting in the labeling $Y_{\text{SkipTag@}K}=[y_{t_1}, \ldots, y_{t_K}]$. We denote $y_t$ as the ground-truth of the $t$-th frame, and $\tilde{y_t}$ is equal to $y_t$ if a ground-truth exists for the $t$-th frame or a previously generated pseudo-label if such was generated for the $t$-th frame.

\subsection{Feature extraction}
\label{subsec:feature-extraction}
Following the approach proposed by Ramesh \etal \cite{ramesh2023dissecting}, we utilize a ResNet-50 \cite{he2016deep} model to extract frame-wise features.  The ResNet-50 model is initialized with weights pre-trained on ImageNet \cite{deng2009imagenet} and then fine-tuned using self-supervised learning on the corresponding dataset with the DINO \cite{caron2021emerging} method.

\subsection{Pseudo-labels generation}
\label{subsec:pseudo-label-generation}
Using the extracted features, we can utilize our temporal model $M(\cdot)$ to generate a probability matrix $P \in {[0, 1]}^{T \times C}$, where $p_t$ is the $t$-th row containing an estimated distribution between phases for the $t$-th frame. We denote $P_{a, b}$ as the predicted probability of the $a$-th frame for the $b$-th phase type.
The predicted phase for the $t$-th frame is determined by selecting the phase with the highest probability according to the estimated distribution. 
Our training method relies on the use of pseudo-labels, which are generated using a technique similar to the Uncertainty-Aware Temporal Diffusion (UATD) method proposed by Ding \etal \cite{ding2023less}. However, our approach differs from UATD by employing entropy as the uncertainty estimator, as opposed to the standard deviation used in \cite{ding2023less}.


\subsection{Loss}
The loss function used in our approach consists of several components, each of which will be explained in detail below.
\\
\textbf{Balanced Classification Loss}, we employ the weighted focal loss \cite{lin2017focal}, which is designed to handle imbalanced data more effectively than the standard cross-entropy loss. This loss component is calculated as follows:
\begin{equation}
    FL(q_o)=-(1-q_o)^{\gamma}log(q_o)
\end{equation}
\begin{equation}
    \mathcal{L}_{cls}=\frac{1}{\underset{t=1}{\overset{T}{\sum}}m_{t}}\underset{t=1}{\overset{T}{\sum}}m_{t}\cdot w_{\tilde{y_{t}}}\cdot FL(P_{t, \tilde{y_t}})
\end{equation}
where $m_{t}=1$ if a label (either ground truth or pseudo-label) exists for the $t$-th frame and 0 otherwise.
$T$ is the total number of frames in the video, $P_{t, \tilde{y_t}}$ is the predicted probability of the model for the $t$-th frame belonging to the label class $\tilde{y_t}$, and $\gamma$ is a hyper-parameter that controls the focus on hard examples.
$w_{\tilde{y_i}}$ is an inverse class frequency weight \cite{lertnattee2004analysis} calculated as follows:
\begin{equation}
    w_{c}=\text{\ensuremath{\frac{N}{N_{c}}}}
\end{equation}
where $N$ is the total number of annotated frames and $N_c$ is the number of annotated frames belonging to the $c$-th phase option. This weighting scheme helps to mitigate the impact of class imbalance during training.

\textbf{Entropy Loss} aims to encourage the model to be more certain about its predictions. It is calculated as follows:
\begin{equation}
    l_{\text{Entropy}}(q)=H(q)=-\underset{j=1}{\overset{C}{\sum}}q_{j}\cdot log(q_{j}) 
    \label{eq:H_entropy}
\end{equation}
\begin{equation}
    \mathcal{L}_{\text{Entropy}}=\frac{1}{\underset{t=1}{\overset{T}{\sum}}m_{t}}\underset{t=1}{\overset{T}{\sum}}m_{t} \cdot l_{\text{Entropy}}(p_{t})\quad 
\end{equation}

\textbf{Confidence loss} is adopted from Li \etal \cite{li2021temporal}.
This loss encourages the model to predict a monotonic increase and decrease in the confidence of a phase around a ground truth prediction. This loss helps to suppress outliers and enforce temporal consistency in the model's predictions.
The loss is calculated as follows:
\begin{equation}
\delta_{t, y_{t_{i}}}=\begin{cases}
\max\left(0,\log(P_{t,y_{t_{i}}})-\log(P_{t-1,y_{t_{i}}})\right) & if\ t\geq t_{i}\\
\max\left(0,\log(P_{t-1,y_{t_{i}}})-\log(P_{t,y_{t_{i}}})\right) & if\ t<t_{i}
\end{cases}
\end{equation}
\begin{equation}
    \mathcal{L}_{conf} = 
    \frac{1}{T^{'}}
    \sum\limits_{i=2}^{\text{TP}-1}
    \left(
    \sum\limits_{t=t_{(i-1)}}^{t_{(i+1)}} \delta_{t, y_{t_i}}
    \right)
\end{equation}

where $T_{sparse}=\{t_1, \ldots, t_{\text{TP}}\}$ is the set of time points of annotated frames and TP is the number of annotated time points. $T^{'} = 2(t_{\text{TP}
} - t_1)$

\textbf{Smoothness loss} is designed to encourage the model to predict a smooth phase segmentation, penalizing changes between neighboring frames. This is done based on a truncated mean squared error over the log probabilities as done in \cite{farha2019ms, li2021temporal ,ding2023less}.
The smoothness loss is calculated as follows: 
\begin{equation}
    \Delta_{t,c}=|log(P_{t,c})-log(P_{t-1,c})|
\end{equation}
\begin{equation}
    \tilde{\Delta}_{t,c}=\begin{cases}
    \Delta_{t,c} & \Delta_{t,c}\leq\tau_{S}\\
    \tau_{S} & otherwise
    \end{cases}
\end{equation}
\begin{equation}
    \mathcal{L}_{S}=\frac{1}{(T - 1) \cdot C}
    \sum\limits_{t=2}^{T}
    \sum\limits_{c=1}^{C}
    \tilde{\Delta}_{t,c}^{2}
\end{equation}

Where $\tau_{S}$ is a threshold hyper-parameter.

\textbf{Star Temporal Classification (STC) loss}, which is a variation of the CTC \cite{graves2006connectionist} loss,   
is used to encourage the model to predict the correct order of the phases.
CTC loss aims to tackle misalignment between input and output sequences of varying lengths, and is widely used for automatic speech recognition \cite{li2022recent} and handwritten text
recognition.
\cite{alkendi2024advancements}.
The CTC introduces an additional $blank$ token
and a collapse function $B(\cdot)$ that maps frame-wise predictions to dense predictions by removing $blank$ tokens and repeating predictions.
Let $X$ be a video containing $T$ frames and a segment-based labeling $y_{\text{seg}}\in \mathcal{P}^{PN}$.

The CTC loss can be expressed as follows:
\begin{align}
    \mathcal{L}_{CTC} 
        &= - \ln \, P(y_{\text{seg}}|X) \\
        &= - \ln \, \sum\limits_{\{\pi | B(\pi) = y_{\text{seg}}\}} P(\pi| X)
\end{align}
where $\pi=[\pi_{1}, \ldots, \pi_{T}]\in \mathcal{P}^T$ represents a sparse phase prediction that would have collapsed to the correct segment-based labeling $y_{seg}$.
Using the CTC's assumption of 
conditional independence
\begin{equation}
    P(\pi| X) = \prod\limits_{t=1}^{T} P(\pi_t|t, X) 
\end{equation}
$P(\pi_{t}|t, X)$ are predicted by the model. 
The CTC loss can be efficiently calculated using dynamic programming.
CTC can also be calculated using weighted finite-state transducers \cite{hannun2020differentiable}.
STC aims to allow learning from weakly supervised labels. The STC is based on adding an additional $star$ token that can represent a missing label. This idea allows handling of a flexible number of missing tokens, while encouraging the model to predict the current phase order.  

To obtain $y_{\text{seg}}$, we use $\tilde{y}=\{\tilde{y_t}:  1 \leq t \leq T \text{ and } \tilde{y_t} \text{ exists}\}$ and remove consecutive duplicate labels.
In experiments where STC is used, we add an additional $blank$ phase that the model can predict, model prediction from a close neighboring frame are copied after the STC loss is calculated to allow correct flow of the method. 


The total loss is a weighted sum of the individual loss components:\begin{equation}
    \mathcal{L} = 
    \mathcal{L}_{cls}
    + \alpha_{1} \mathcal{L}_{S}
    + \alpha_{2} \mathcal{L}_{\text{Entropy}}
    + \alpha_{3} \mathcal{L}_{conf}
    + \alpha_{4} \mathcal{L}_{stc}
\end{equation}
where $\alpha_{1}, 
       \alpha_{2},
       \alpha_{3},
       \text{and } \alpha_{4}$ are hyper-parameters.
\subsection{Additional training stage}
\begin{figure}
    \centering
    \includegraphics[width=\columnwidth]{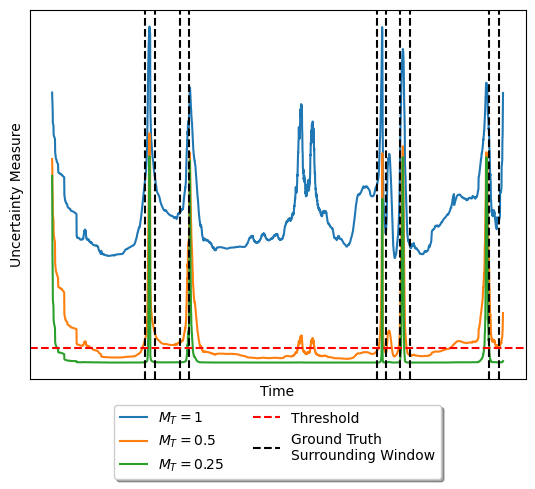}

    \caption{Illustration of the uncertainty measure used to identify phase transition events with different temperature scaling values (T) on a Cholec80 video. The black vertical lines represent a surrounding window of 2W frames centered on the ground truth transition events between surgical phases. Lower temperature values result in more stable uncertainty measure, enabling more robust transition event detection compared to using no temperature scaling (T=1). The red horizontal line represent the uncertainty threshold.
    }
    \label{fig:unceranity-graph}
\end{figure}

Using the trained model $M(\cdot)$, we create fixed partial pseudo-labels that are then used to train a new model.

When generating those pseudo-labels, we aim to utilize all of the model's predictions, except for those corresponding to transition moments and their surrounding frames.  To achieve this, we first need to detect transition moments. We used a scaled entropy measure for this
\begin{equation}
    U(l)=H\left(\operatorname{softmax}\left(\frac{l}{M_T}\right)\right)
\end{equation}
where $l$ is the model logits's output, $M_T$ is the scaling temperature and $H$ is the entropy measure as defined in the Eq. \ref{eq:H_entropy}. We use $0<M_T<1$ to push the model's outputs to the far ends of the spectrum. By employing this estimator, we can easily detect the transition events using a simple threshold $\tau_{transition}$ as illustrated in Fig. \ref{fig:unceranity-graph}. We now consider all predictions that are not within a window of $W$ frames from a phase transition event and utilize them as pseudo-labels for further training.
The new model is trained using an unweighted focal loss combined with a smoothness loss component.

\subsection{Implementation Details}
All experiments were performed on NVIDIA A100 Tensor Core and NVIDIA RTX 6000 Ada Generation GPUS.  Both datasets were down-sampled to 1fps. We follow Ramesh \etal \cite{ramesh2023dissecting} regarding the self-supervised feature extractor fine-tunning configuration. We used the Adam optimizer with initial learning rate of 5e-4 for training the temporal model for 50 epochs. We resize the images to a 224 × 399 resolution before propagating through our feature extractor. We employ the TCN-based model suggested by Li \etal \cite{li2021temporal} as our temporal model. We use $W=25, \tau_{S}=16$ for all experiments.

\section{Experiments}
\label{sec:experiments}
\subsection{Datasets}
\noindent \textbf{Cholec80} \cite{twinanda2016endonet} contains 80 videos of laparoscopic cholecystectomy procedures, with an average duration of 38 minutes. The dataset contain 7 types of phases. It was recorded at 25 fps with a resolution of 854x840 or 1920x1080. We follow Ding \etal \cite{ding2023less} and use the first 40 videos as a training set and use the rest of the videos as a test set.
\\
\textbf{MultiBypass140} \cite{lavanchy2023challenges}
contains 140 videos of laparoscopic Roux-en-Y gastric bypass surgeries, with an average duration of 91 minutes. The dataset contains 12 types of phases. It was recorded at 25 fps in two medical centers with a resolution of 720x576 or 854x480 or 1920x1080.
we follow Lavanchy and Ramesh \etal \cite{lavanchy2023challenges} and split the dataset to three parts, training, validation and test, which contain 80, 20, and 40 videos respectively. 
\subsection{Evaluation Metrics}
We follow previus works  \cite{shi2021semi, ding2023less, jin2017sv} and report frame-level evaluation metrics,  accuracy (AC), precision (PR), recall (RE), Jaccard (JA), and F1. For each phase prediction P and a ground truth GT , PR, RE, JA, and F1 are calculated as follows:
\begin{equation}
\begin{split}
PR = \frac{|GT \cap P|}{|P|}, 
RE = \frac{|GT \cap P|}{|GT|}, \\
JA = \frac{|GT\cap P|}{|GT \cup P|}, 
F1 = \frac{2 \cdot PR \cdot RE}{RE + PR}
\end{split}
\end{equation}
the scores for each phase are averaged across all the phases in a video. Accuracy is calculated globally for each video. When evaluating on Cholec80, we follow \cite{ramesh2023dissecting, ding2023less} and report 10-second 'relaxed' metrics, meaning that we allow two correct phases in the 10 second surrounding window of each phase transition. 

\subsection{Sampling Distribution}
Figure \ref{fig:label-histogram} presents the distribution of per-frame phase annotations in the Cholec80 dataset. The full distribution, shown in yellow, illustrates a significant imbalance between classes, with some phases occurring much more frequently than others. This imbalance poses a challenge for surgical phase recognition models, as they may struggle to accurately classify underrepresented phases.
In contrast, the timestamp labels, depicted in pink, are distributed more evenly across the phases. This suggests that while the duration of phases may vary, the number of phase occurrences is relatively balanced in the dataset. The more uniform distribution of timestamp labels can be attributed to the fact that they are selected based on the presence of each phase rather than their duration.
In order to create SkipTag@K annotations, we split the video into K equal partitions and sample uniformly from each partition a single sample. 
The blue shades represent the SkipTag@K sampling method with K values of 2, 4, and 7, where K denotes the fixed number of frames annotated per video. As K increases, the SkipTag@K distributions more closely resemble the full distribution, indicating that this sampling strategy effectively captures the original data distribution. The similarity between the SkipTag@7 and full distributions suggests that annotating just 7 frames per video can provide a representative sample of the phase distribution in the Cholec80 dataset. 

\begin{figure}
    \centering
    \includegraphics[width=\columnwidth]{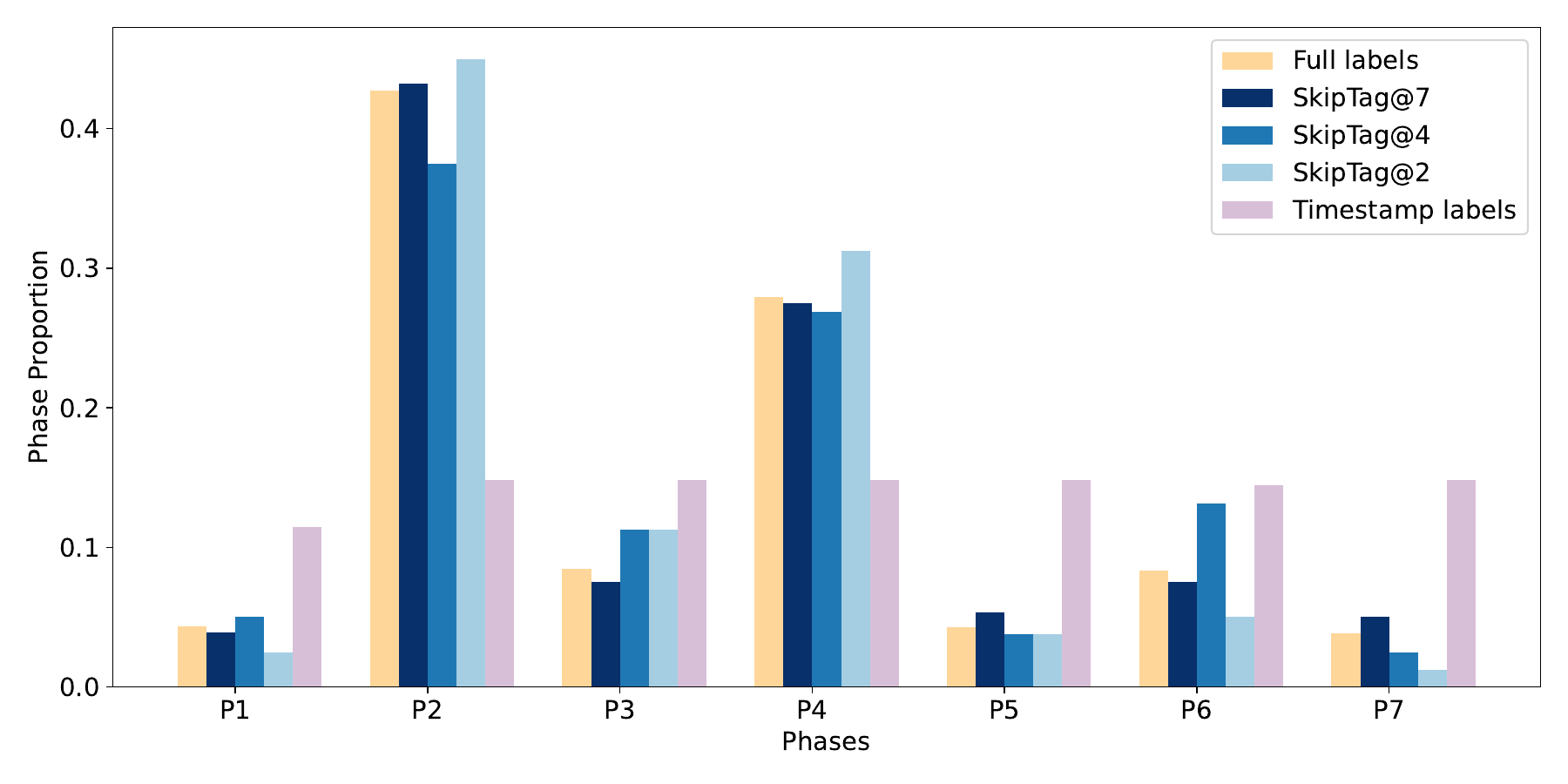}
    \caption{Distribution of per-frame phase annotations in the Cholec80 dataset. The full distribution (yellow) reveals significant class imbalance, while the timestamp labels (pink) are more evenly distributed. SkipTag@K sampling with K=2, 4, and 7 (blue shades) effectively captures the original data distribution, with increasing similarity to the full distribution as K increases.}
    \label{fig:label-histogram}
\end{figure}
\subsection{Robustness to Missing Phase Annotations}




\begin{table*}
    \centering
    \begin{adjustbox}{width=\textwidth}
    \begin{tabular}{|c|c|c|c|c|c|c|c|c|c|c|}
\hline 
\multirow{2}{*}{Method} & \multicolumn{5}{c|}{Cholec80} & \multicolumn{5}{c|}{MultiBypass140}\tabularnewline
 & \multicolumn{1}{c}{RE (\%)} & \multicolumn{1}{c}{PR (\%)} & \multicolumn{1}{c}{JA (\%)} & \multicolumn{1}{c}{AC (\%)} & F1 (\%) & \multicolumn{1}{c}{RE (\%)} & \multicolumn{1}{c}{PR (\%)} & \multicolumn{1}{c}{JA (\%)} & \multicolumn{1}{c}{AC (\%)} & F1 (\%)\tabularnewline
\hline 
\multicolumn{11}{c}{Timestamp supervision (miss rate = 0)}\tabularnewline
\hline 
Ding \etal \cite{ding2023less} & 90.5$\pm$5.9 & 89.5$\pm$4.4 & \textbf{79.9$\pm$8.5} & \textbf{91.9$\pm$5.6} & - & 74.3$\pm$10.0 & 75.7$\pm$13.2 & 65.2$\pm$11.6 & 87.2$\pm$10.6 & 72.1$\pm$11.3\tabularnewline
Ours & \textbf{97$\pm$9.8} & 84.8$\pm$6.9 & 76.1$\pm$8.8 & 90.4$\pm$5.9 & 87.9$\pm$5.4 & \textbf{78.3$\pm$11.7} & \textbf{75.9$\pm$11.5} & \textbf{67.8$\pm$13} & \textbf{88.4$\pm$9.7} & \textbf{74.6$\pm$12.3}\tabularnewline
\hline 
\multicolumn{11}{c}{Miss rate = 0.1}\tabularnewline
\hline 
Ding \etal \cite{ding2023less} & 86.4$\pm$11.5 & \textbf{87.9$\pm$6.7} & 74.9$\pm$8.0 & 86.50$\text{\ensuremath{\pm}}$9.6 & 86.5$\pm$6.7 & 75.1$\pm$10.1 & \textbf{75.2$\pm$12.6} & 65.1$\pm$12.0 & 86.3$\pm$10.0 & 72.1$\pm$11.3\tabularnewline
Ours & \textbf{97.3$\pm$9.3} & 84.1$\pm$6.8 & \textbf{75.4$\pm$8.4} & \textbf{89.8$\pm$5.9} & \textbf{87.4$\pm$5.8} & \textbf{76.9$\pm$11.3} & 73.9$\pm$12.4 & \textbf{65.9$\pm$13.7} & \textbf{87.5$\pm$9.9} & \textbf{72.8$\pm$13}\tabularnewline
\hline 
\multicolumn{11}{c}{Miss rate = 0.2}\tabularnewline
\hline 
Ding \etal \cite{ding2023less} & 81.8$\pm$19.2 & \textbf{85.2$\pm$8.45} & 67.8$\pm$14.0 & 84$\pm$8.8 & 81.6$\pm$8.45 & 70.5$\pm$13.0 & 69.3$\pm$13.9 & 58.5$\pm$12.6 & 82.5$\pm$10.3 & 66.4$\pm$13.1\tabularnewline
Ours & \textbf{96.8$\pm$9.3} & 81.9$\pm$7.2 & \textbf{72.7$\pm$9.15} & \textbf{86.9$\pm$7.4} & \textbf{84.9$\pm$6.5} & \textbf{75.7$\pm$12.2} & \textbf{73.1$\pm$12.3} & \textbf{64.8$\pm$13.6} & \textbf{87$\pm$9.8} & \textbf{71.7$\pm$13}\tabularnewline
\hline 
\multicolumn{11}{c}{Miss rate = 0.3}\tabularnewline
\hline 
Ding \etal \cite{ding2023less} & 76.6$\pm$16.1 & 80.2$\pm$9.6 & 58.9$\pm$13.5 & 78$\pm$12.1 & 77.1$\pm$9.6 & 61.4$\pm$11.3 & 60.6$\pm$12.2 & 46.7$\pm$11.3 & 71.5$\pm$14.3 & 55.7$\pm$11.5\tabularnewline
Ours & \textbf{96.5$\pm$9.5} & \textbf{82.4$\pm$7.4} & \textbf{73.2$\pm$9} & \textbf{87.4$\pm$7.5} & \textbf{85.5$\pm$5.8} & \textbf{76.7$\pm$12.4} & \textbf{72.2$\pm$13} & \textbf{63.7$\pm$14.6} & \textbf{85.7$\pm$9.5} & \textbf{71.1$\pm$14}\tabularnewline
\hline 
\end{tabular}
    \end{adjustbox}
    \caption{Robustness to missing phase annotations}
    \label{tab:all_missing}
\end{table*}

    

To demonstrate the impact of missing phase annotations and the robustness of our method, 
we compare our approach with Ding \etal's method. 
We simulated missing phase annotations by randomly removing phase labels from the timestamp annotations with 
varying miss rate probabilities $p_m$. 
As shown in Table \ref{tab:all_missing}, Ding \etal's method is significantly affected by missing phase annotations. 
This is evident from the increased standard deviations in the evaluation metrics on the Cholec80 dataset 
and the substantial drop in performance, ranging from a 10 \% to a 28 \% decline when comparing the timestamp results 
($p_m=0$) to the results with $p_m=0.3$. 
In contrast, our method's performance only slightly declines, 
achieving competitive results on the Cholec80 dataset in the timestamp setup, 
surpassing Ding \etal's method in the timestamp setup, 
and consistently outperforming their approach on both datasets across all missing rate settings and almost all metrics.

Figure \ref{fig:label-histogram} further illustrates the performance comparison between our method and Ding \etal's approach under different missing rate probabilities.

\begin{figure*}[!ht]
    \centering
    \includegraphics[width=0.8\textwidth]{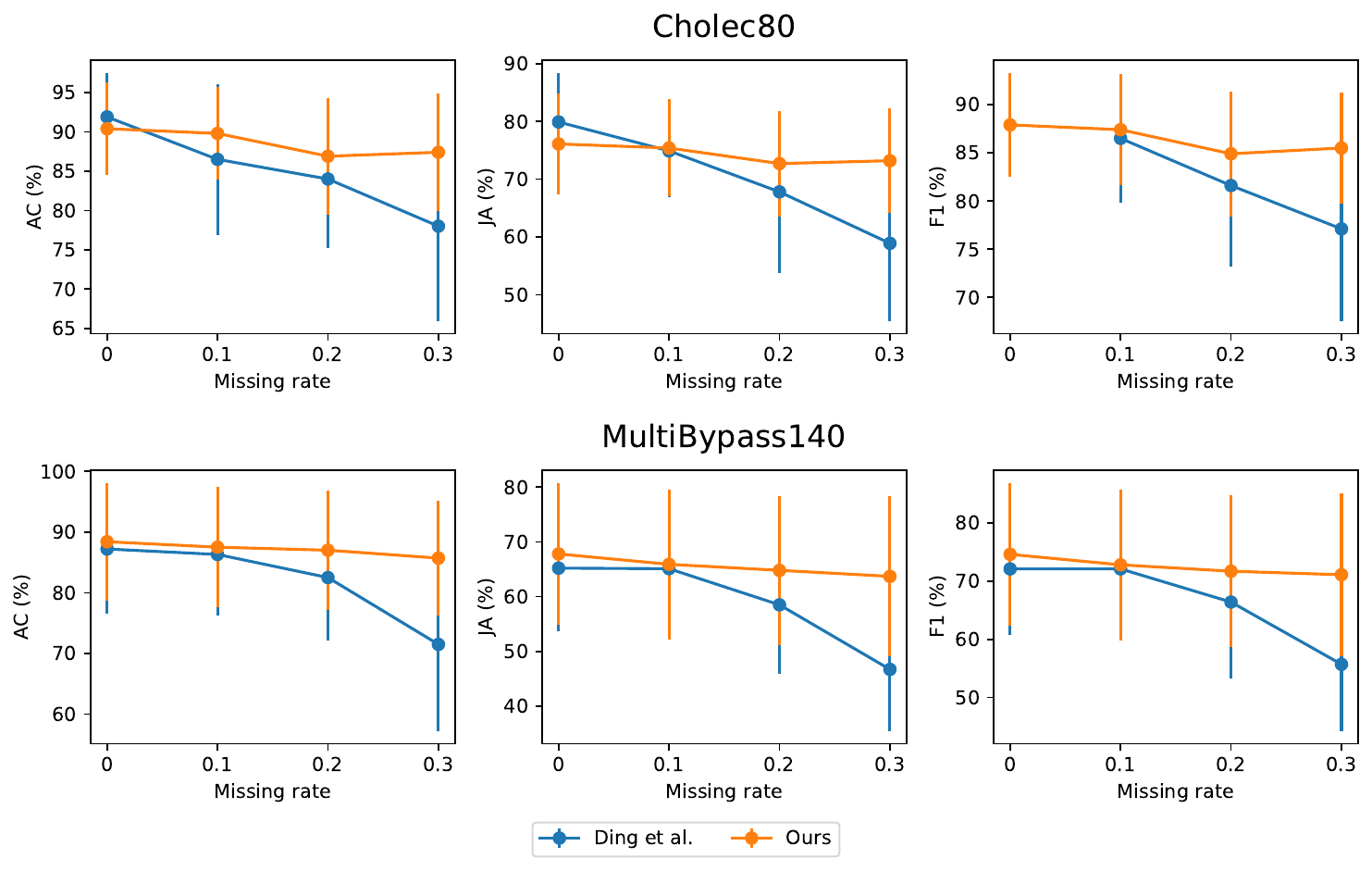}
    \caption{Robustness comparison between our method and Ding \etal's approach under different missing rate probabilities for the Cholec80 and MultiBypass140 datasets. The six subfigures depict the evaluation metrics of accuracy, Jaccard index, and F1 score for both datasets. As the missing rate increases, Ding \etal's method experiences a significant drop in performance across all metrics on both datasets, while our method maintains stable performance with only a slight decline.}
    \label{fig:label-histogram}
\end{figure*}

\subsection{SkipTag@K Evaluation}
In this section, we evaluate the performance of our model in the challenging SkipTag@K setup, where only K frames from each video are annotated. We present results for three K values relative to the average number of phases $N_{avg}$ in the corresponding dataset. The K values are $round(N_{avg}), \ceil{\frac{N_{avg}}{2}},  \text{and} \ceil{\frac{N_{avg}}{4}}$, where $N_{avg}$ is 6.8 for Cholec80 and 9.3 for MultiBypass140.

\begin{table*}[!ht]
\centering
    \begin{tabular}{llcccccc}
        \toprule
        Setup & Method & RE (\%) & PR (\%) & JA (\%) & AC (\%) & F1 (\%) \\
        \midrule
        \multirow{2}{*}{SkipTag@7} 
        & Ding \etal \cite{ding2023less} & 86.5$\pm$11.5 & 87.9$\pm$6.7 & \textbf{74.9$\pm$8.0} & 86.5$\pm$9.6 & \textbf{86.5$\pm$6.7} \\
        & Ours & \textbf{87.8$\pm$11.5} & \textbf{88.1$\pm$8.5} & 73.6$\pm$11.9 & \textbf{91.2$\pm$5.9} & 86.4$\pm$7.8 \\
        \midrule
        \multirow{2}{*}{SkipTag@4} 
        & Ding \etal \cite{ding2023less} & 50.4$\pm$32.0 & \textbf{84.4$\pm$13.3} & 40.6$\pm$24.7 & 76.7$\pm$9.8 & 57.1$\pm$13.3 \\
        & Ours & \textbf{84.0$\pm$13.0} & 82.3$\pm$10.4 & \textbf{66.9$\pm$13.5} & \textbf{88.9$\pm$7.6} & \textbf{84.1$\pm$9.7} \\
        \midrule
        \multirow{2}{*}{SkipTag@2} 
        & Ding \etal \cite{ding2023less} & 32.1$\pm$39.3 & \textbf{88.0$\pm$17.7} & 23.8$\pm$26.0 & 65.6$\pm$12.1 & 39.1$\pm$17.7 \\
        & Ours & \textbf{71.2$\pm$12.2} & 81.0$\pm$9.0 & \textbf{57.9$\pm$11.1} & \textbf{83.6$\pm$7.1} & \textbf{78.1$\pm$7.9} \\
        \bottomrule
    \end{tabular}
\caption{SkipTag@K results on the Cholec80 dataset}
\label{tab:cholec80-SkipTag-test}
\end{table*}
\begin{table*}
\centering
    \begin{tabular}{llcccccc}
        \toprule
        Setup & Method & RE (\%) & PR (\%) & JA (\%) & AC (\%) & F1 (\%) \\
        \midrule
        \multirow{2}{*}{SkipTag@9} 
        & Ding \etal \cite{ding2023less} & 46.6$\pm$7.3 & 44.1$\pm$10.1 & 37.9$\pm$9.4 & 76.5$\pm$11.3 & 42.8$\pm$8.9 \\
        & Ours & \textbf{77.8$\pm$10.5} & \textbf{75.8$\pm$10.6} & \textbf{68.3$\pm$12.3} & \textbf{88.7$\pm$10.5} & \textbf{74.7$\pm$11.2} \\
        \midrule
        \multirow{2}{*}{SkipTag@5} 
        & Ding \etal \cite{ding2023less} & 31.5$\pm$5.8 & 26.8$\pm$8.3 & 22.2$\pm$6.8 & 64.4$\pm$11.7 & 26.8$\pm$6.8 \\
        & Ours & \textbf{74.9$\pm$11.9} & \textbf{75.0$\pm$13.3} & \textbf{66.3$\pm$14.3} & \textbf{87.7$\pm$11.7} & \textbf{72.7$\pm$13.4} \\
        \midrule
        \multirow{2}{*}{SkipTag@3} 
        & Ding \etal \cite{ding2023less} & 25.0$\pm$8.4 & 21.5$\pm$11.2 & 16.1$\pm$8.3 & 49.9$\pm$16.1 & 20.3$\pm$9.1 \\
        & Ours & \textbf{68.6$\pm$12.8} & \textbf{65.7$\pm$14.4} & \textbf{58.5$\pm$14.4} & \textbf{85.1$\pm$12.3} & \textbf{65.0$\pm$14.3} \\
        \bottomrule
    \end{tabular}
\caption{SkipTag@K results on the MultiBypass140 dataset}
\label{tab:MultiBypass140-SkipTag-test}
\end{table*}

Tables \ref{tab:cholec80-SkipTag-test} and \ref{tab:MultiBypass140-SkipTag-test} demonstrate that our model achieves impressive results in the SkipTag@K setup, despite the limited number of annotated frames. 
Remarkably, the decline in performance when reducing the number of annotated samples is relatively small compared to the reduction in the number of samples itself. For instance, on both datasets, the F1 measure drops by less than 13\% when comparing the results obtained with the largest tested K value to those obtained with the lowest tested K value, even though the larger number of samples is at least 3 times as large. Similarly, the accuracy drops by less than 9\% under the same conditions.
In contrast, Ding \etal's method experiences a drastic performance drop when the number of annotated samples is reduced.
 
\subsection{Ablations}

\subsubsection{Component Analysis}

\begin{table*}
    \centering
        \begin{tabular}{lcccccc}
        \toprule
        Method & RE (\%) & PR (\%) & JA (\%) & AC (\%) & F1 (\%) \\
        \midrule
        Base & 34.1$\pm$17.5 & 54.3$\pm$15.8 & 22.6$\pm$15.2 & 50.9$\pm$17.1 & 58$\pm$17.8 \\
        +DINO FE & 28.3$\pm$2.7 & 67.3$\pm$10.6 & 18.7$\pm$4.0 & 67.7$\pm$10.9 & 72.1$\pm$14.9 \\
        +Conf loss & 37.8$\pm$7.3 & 77.4$\pm$9.5 & 27.9$\pm$6.2 & 70.9$\pm$9.5 & 78$\pm$11.0 \\
        +Focal loss & 63$\pm$11.0 & 76.6$\pm$12.4 & 51.4$\pm$10.7 & 81.8$\pm$9.1 & 77.1$\pm$8.8 \\
        +Loss reweighting & 81.5$\pm$10.0 & 86.8$\pm$9.4 & 68.1$\pm$9.9 & 89.2$\pm$6.6 & 84.6$\pm$6.7 \\
        +STC loss & 86.8$\pm$10.6 & \textbf{88.1$\pm$7.2} & 72.5$\pm$11.0 & 90.8$\pm$5.6 & 85.6$\pm$6.9 \\
        +Additional training & \textbf{87.8$\pm$11.5} & \textbf{88.1$\pm$8.5} & \textbf{73.6$\pm$11.9} & \textbf{91.2$\pm$5.9} & \textbf{86.4$\pm$7.8} \\
        \bottomrule
    \end{tabular}
    \caption{Ablation study demonstrating the impact of each component in our method, using the SkipTag@7 setup on the Cholec80 dataset.}
    \label{tab:cholec80_ablation}
\end{table*}

Table \ref{tab:cholec80_ablation} presents an ablation study that highlights the contribution of each component in our method, using the SkipTag@7 setup on the Cholec80 dataset. The first modification is the replacement of the feature extractor from a ResNet-50 model pretrained on ImageNet to a self-supervised DINO model fine-tuned on the relevant dataset. This change leads to significant improvements across most metrics and reduces the model's standard deviation, indicating more robust performance.
Subsequent modifications demonstrate the impact of various loss functions and training strategies. The introduction of the confidence loss, focal loss, loss re-weighting, and STC loss all contribute to incremental improvements in the model's performance.
The final component is the additional training phase, which utilizes the generated pseudo-labels to further refine the model. This step results in further performance gains, achieving the highest scores across all evaluation metrics.




\subsubsection{Additional training stage}
\label{subsec: addtional-traininig}

\begin{table*}
    \centering
    \begin{tabular}{llccccc}
        \toprule
        & & RE (\%) & PR (\%) & JA (\%) & AC (\%) & F1 (\%) \\
        \midrule
        \multirow{3}{*}{Timestamp supervision} 
        & Base & \textbf{97.5$\pm$9.2} & \textbf{84.9$\pm$7.6} & \textbf{76.4$\pm$10.3} & 90.3$\pm$6.8 & 87.9$\pm$6.9 \\
        & $M_{T}$=0.5 & 97.1$\pm$9.6 & 82.9$\pm$7.4 & 73.9$\pm$9.3 & 88.7$\pm$6.2 & 85.9$\pm$6.7 \\
        & $M_{T}$=0.25 & 97.0$\pm$9.8 & 84.8$\pm$6.9 & 76.1$\pm$8.8 & \textbf{90.4$\pm$5.9} & 87.9$\pm$5.4 \\
        \midrule
        \multirow{3}{*}{Missing 0.1} 
        & Base & 96.5$\pm$9.1 & 82.5$\pm$7.4 & 73.0$\pm$9.2 & 87.8$\pm$6.7 & 85.4$\pm$6.6 \\
        & $M_{T}$=0.5 & 96.6$\pm$9.6 & 82.7$\pm$6.9 & 73.5$\pm$8.6 & 88.3$\pm$6.2 & 85.5$\pm$6.2 \\
        & $M_{T}$=0.25 & \textbf{97.3$\pm$9.3} & \textbf{84.1$\pm$6.8} & \textbf{75.4$\pm$8.4} & \textbf{89.8$\pm$5.9} & \textbf{87.4$\pm$5.8} \\
        \midrule
        \multirow{3}{*}{Missing 0.2} 
        & Base & \textbf{96.8$\pm$9.3} & 81.9$\pm$7.2 & 72.7$\pm$9.1 & 86.9$\pm$7.4 & 84.9$\pm$6.5 \\
        & $M_{T}$=0.5 & 96.7$\pm$9.1 & 82.0$\pm$6.6 & 72.5$\pm$8.3 & 87.3$\pm$6.2 & 85.0$\pm$5.6 \\
        & $M_{T}$=0.25 & 96.6$\pm$9.1 & \textbf{83.2$\pm$6.2} & \textbf{74.0$\pm$8.0} & \textbf{88.0$\pm$6.6} & \textbf{85.8$\pm$5.3} \\
        \midrule
        \multirow{3}{*}{Missing 0.3} 
        & Base & \textbf{96.7$\pm$9.3} & 80.7$\pm$7.9 & 71.3$\pm$10.4 & 85.4$\pm$8.9 & 83.2$\pm$8.7 \\
        & $M_{T}$=0.5 & 96.1$\pm$9.6 & 80.9$\pm$7.1 & 71.5$\pm$9.4 & 86.5$\pm$7.2 & 84.3$\pm$6.1 \\
        & $M_{T}$=0.25 & 96.5$\pm$9.5 & \textbf{82.4$\pm$7.4} & \textbf{73.2$\pm$9.0} & \textbf{87.4$\pm$7.5} & \textbf{85.5$\pm$5.8} \\
        \midrule
        \multirow{3}{*}{SkipTag@7} 
        & Base & 86.8$\pm$10.6 & 88.1$\pm$7.2 & 72.5$\pm$11.0 & 90.8$\pm$5.6 & 85.6$\pm$6.9 \\
        & $M_{T}$=0.5 & 83.5$\pm$12.7 & 83.8$\pm$10.7 & 69.2$\pm$13.3 & 88.4$\pm$8.2 & 84.7$\pm$8.1 \\
        & $M_{T}$=0.25 & \textbf{87.8$\pm$11.5} & 88.1$\pm$8.5 & \textbf{73.6$\pm$11.9} & \textbf{91.2$\pm$5.9} & \textbf{86.4$\pm$7.8} \\
        \midrule
        \multirow{3}{*}{SkipTag@4} 
        & Base & 83.0$\pm$12.1 & \textbf{84.0$\pm$7.7} & 66.0$\pm$11.8 & 88.3$\pm$6.3 & 80.7$\pm$9.9 \\
        & $M_{T}$=0.5 & 53.9$\pm$10.5 & 76.3$\pm$11.9 & 40.6$\pm$9.1 & 80.1$\pm$7.1 & 81.5$\pm$12.1 \\
        & $M_{T}$=0.25 & \textbf{84.0$\pm$13.0} & 82.3$\pm$10.4 & \textbf{66.9$\pm$13.5} & \textbf{88.9$\pm$7.6} & \textbf{84.1$\pm$9.7} \\
        \midrule
        \multirow{3}{*}{SkipTag@2} 
        & Base & \textbf{73.8$\pm$12.5} & \textbf{81.6$\pm$9.4} & \textbf{59.4$\pm$12.7} & 83.4$\pm$7.8 & 77.4$\pm$9.1 \\
        & $M_{T}$=0.5 & 69.8$\pm$11.9 & 78.1$\pm$10.3 & 54.9$\pm$11.5 & 81.4$\pm$8.2 & 74.8$\pm$9.2 \\
        & $M_{T}$=0.25 & 71.2$\pm$12.2 & 81.0$\pm$9.0 & 57.9$\pm$11.1 & \textbf{83.6$\pm$7.1} & \textbf{78.1$\pm$7.9} \\
        \bottomrule
    \end{tabular}
    \caption{Impact of temperature scaling on the additional training stage for the Cholec80 dataset.}
    \label{tab:cholec80-additonal-training-stage}
\end{table*}

\begin{table*}
    \centering
        \begin{tabular}{llccccc}
        \toprule
        & & RE (\%) & PR (\%) & JA (\%) & AC (\%) & F1 (\%) \\
        \midrule
        \multirow{3}{*}{Timestamp supervision} 
        & Base & 72.6$\pm$10.0 & 70.0$\pm$10.6 & 61.9$\pm$12.0 & 86.7$\pm$10.0 & 69.0$\pm$11.3 \\
        & $M_{T}$=0.5 & \textbf{78.3$\pm$11.7} & 75.9$\pm$11.5 & \textbf{67.8$\pm$13.0} & \textbf{88.4$\pm$9.7} & \textbf{74.6$\pm$12.3} \\
        & $M_{T}$=0.25 & 77.2$\pm$12.0 & 75.9$\pm$12.2 & 67.2$\pm$13.2 & 88.1$\pm$9.8 & 73.9$\pm$12.8 \\
        \midrule
        \multirow{3}{*}{Missing 0.1} 
        & Base & 71.3$\pm$10.5 & 69.0$\pm$10.6 & 60.6$\pm$12.0 & 85.9$\pm$10.0 & 67.8$\pm$11.2 \\
        & $M_{T}$=0.5 & 76.9$\pm$11.3 & 73.9$\pm$12.4 & 65.9$\pm$13.7 & 87.5$\pm$9.9 & 72.8$\pm$13.0 \\
        & $M_{T}$=0.25 & \textbf{78.2$\pm$11.4} & \textbf{75.3$\pm$13.2} & \textbf{67.2$\pm$14.0} & \textbf{87.7$\pm$10.3} & \textbf{74.1$\pm$13.3} \\
        \midrule
        \multirow{3}{*}{Missing 0.2} 
        & Base & 73.9$\pm$12.3 & 72.1$\pm$12.4 & 62.7$\pm$13.4 & 85.4$\pm$10.7 & 70.1$\pm$13.0 \\
        & $M_{T}$=0.5 & \textbf{75.7$\pm$12.2} & \textbf{73.1$\pm$12.3} & \textbf{64.8$\pm$13.6} & \textbf{87.0$\pm$9.8} & \textbf{71.7$\pm$13.0} \\
        & $M_{T}$=0.25 & 71.1$\pm$11.9 & 72.1$\pm$13.3 & 62.0$\pm$13.6 & 85.1$\pm$12.6 & 68.9$\pm$13.0 \\
        \midrule
        \multirow{3}{*}{Missing 0.3} 
        & Base & 75.0$\pm$11.1 & 71.2$\pm$12.0 & 62.0$\pm$12.6 & 84.7$\pm$10.1 & 70.1$\pm$12.1 \\
        & $M_{T}$=0.5 & 76.7$\pm$12.4 & 72.2$\pm$13.0 & 63.7$\pm$14.6 & 85.7$\pm$9.5 & 71.1$\pm$14.0 \\
        & $M_{T}$=0.25 & 76.7$\pm$12.0 & \textbf{72.5$\pm$13.0} & \textbf{63.8$\pm$14.3} & \textbf{85.8$\pm$10.0} & \textbf{71.3$\pm$13.7} \\
        \midrule
        \multirow{3}{*}{SkipTag@9} 
        & Base & 72.8$\pm$10.7 & 73.4$\pm$11.5 & 63.9$\pm$13.0 & 87.6$\pm$10.4 & 70.5$\pm$12.2 \\
        & $M_{T}$=0.5 & \textbf{77.8$\pm$10.5} & \textbf{75.8$\pm$10.6} & \textbf{68.3$\pm$12.3} & 88.7$\pm$10.5 & \textbf{74.7$\pm$11.2} \\
        & $M_{T}$=0.25 & 77.7$\pm$10.3 & 75.7$\pm$10.8 & 68.1$\pm$12.1 & 88.7$\pm$10.5 & 74.5$\pm$11.1 \\
        \midrule
        \multirow{3}{*}{SkipTag@5} 
        & Base & 69.3$\pm$12.4 & 73.0$\pm$12.5 & 60.7$\pm$13.8 & 86.1$\pm$11.5 & 67.9$\pm$12.9 \\
        & $M_{T}$=0.5 & \textbf{74.9$\pm$11.9} & 75.0$\pm$13.3 & \textbf{66.3$\pm$14.3} & \textbf{87.7$\pm$11.7} & \textbf{72.7$\pm$13.4} \\
        & $M_{T}$=0.25 & 73.9$\pm$11.9 & \textbf{75.7$\pm$13.3} & 65.5$\pm$14.5 & 87.6$\pm$11.9 & 72.1$\pm$13.5 \\
        \midrule
        \multirow{3}{*}{SkipTag@3} 
        & Base & 62.7$\pm$11.6 & 62.2$\pm$13.8 & 52.7$\pm$13.3 & 83.1$\pm$11.8 & 59.2$\pm$12.9 \\
        & $M_{T}$=0.5 & \textbf{68.6$\pm$12.8} & \textbf{65.7$\pm$14.4} & \textbf{58.5$\pm$14.4} & \textbf{85.1$\pm$12.3} & \textbf{65.0$\pm$14.3} \\
        & $M_{T}$=0.25 & 67.6$\pm$12.8 & 65.4$\pm$14.3 & 57.3$\pm$14.9 & 84.5$\pm$11.8 & 63.8$\pm$14.6 \\
        \bottomrule
    \end{tabular}
    \caption{Impact of temperature scaling on the additional training stage for the MultiBypass140 dataset.}
    \label{tab:MultiBypass140-additonal-training-stage}
\end{table*}

In this section we focus on the pseudo-lebel generation mechanism and investigate the effect of the temperature $M_T$ used in the pseudo-generation mechanism. Tables \ref{tab:cholec80-additonal-training-stage} and \ref{tab:MultiBypass140-additonal-training-stage} present a comparison between the base model (before pseudo-label generation) and models trained on pseudo-labels generated with $M_{T}=0.25$ and $M_{T}=0.5$. The comparison is performed for various setups, including timestamp supervision, missing annotations, and SkipTag@K. 
As illustrated in Figure \ref{fig:unceranity-graph}, lower temperature values result in more extreme uncertainty estimates, leading to a larger number of frames falling below the uncertainty threshold and consequently being annotated with pseudo-labels.
For the Cholec80 dataset (Table \ref{tab:cholec80-additonal-training-stage}), we observe that setting $M_{T}=0.25$ is preferred over setting $M_{T}=0.5$ across all supervision setups. The additional training stage improves the results in most setup. 

For the MultiBypass140 dataset (Table \ref{tab:MultiBypass140-additonal-training-stage}), the optimal temperature value varies depending on the setup. However, both $M_{T}=0.25$ and $M_{T}=0.5$ consistently outperform the base model, demonstrating the effectiveness of the additional training stage across different setups. 

\section{Discussion and Conclusions}
\label{sec:discussion}
In this study, we focused on annotation time efficient techniques for surgical phase recognition.
We addressed the challenge of missing phase annotations in the promising timestamp supervision approach and presented a robust method that is able to tackle this hurdle. This direction paves the way for timestamp annotation in a realistic setup where some phase annotations may be unintentionally omitted, further promoting its adoption as a viable alternative to the fully supervised setup.
Additionally, we introduced SkipTag@K to the surgical domain, offering a flexible trade-off between annotation effort and model performance and demonstrated the competitive results obtained with it.
Remarkably, we achieved impressive performance on both datasets using as few as 2 samples 
for Cholec80 and 3 samples for MultiBypass140 per video
which corresponds to only 0.037\% and 0.0023\% of the corresponding training sets respectively.

While we successfully tackled the issue of missing annotation labels, our work did not address the problem of incorrect label assignments during the annotation process. This opens up an avenue for future research to investigate methods that can handle both missing and incorrect labels. 
In the SkipTag@K experiments, we employed a simple strategy of uniformly selecting a sample from each of the K equally divided video segments. Although this approach yielded impressive results, exploring more sophisticated sampling, as clustering-based techniques, could potentially lead to even better performance. 
Another potential future direction is to investigate the use of easily obtainable weak signals from the surgical procedure itself to efficiently achieve partial labeling. Such signals could include the use of specific surgical tools, changes in the surgical scene, or even audio cues from the operating room. By leveraging these inherent weak signals, we may be able to further reduce the annotation burden while maintaining high performance in surgical phase recognition.




\bibliographystyle{IEEEtran}
\bibliography{references}

\begin{thebibliography}{10}
\providecommand{\url}[1]{#1}
\csname url@samestyle\endcsname
\providecommand{\newblock}{\relax}
\providecommand{\bibinfo}[2]{#2}
\providecommand{\BIBentrySTDinterwordspacing}{\spaceskip=0pt\relax}
\providecommand{\BIBentryALTinterwordstretchfactor}{4}
\providecommand{\BIBentryALTinterwordspacing}{\spaceskip=\fontdimen2\font plus
\BIBentryALTinterwordstretchfactor\fontdimen3\font minus \fontdimen4\font\relax}
\providecommand{\BIBforeignlanguage}[2]{{%
\expandafter\ifx\csname l@#1\endcsname\relax
\typeout{** WARNING: IEEEtran.bst: No hyphenation pattern has been}%
\typeout{** loaded for the language `#1'. Using the pattern for}%
\typeout{** the default language instead.}%
\else
\language=\csname l@#1\endcsname
\fi
#2}}
\providecommand{\BIBdecl}{\relax}
\BIBdecl

\bibitem{maier2017surgical}
L.~Maier-Hein, S.~S. Vedula, S.~Speidel, N.~Navab, R.~Kikinis, A.~Park, M.~Eisenmann, H.~Feussner, G.~Forestier, S.~Giannarou \emph{et~al.}, ``Surgical data science for next-generation interventions,'' \emph{Nature Biomedical Engineering}, vol.~1, no.~9, pp. 691--696, 2017.

\bibitem{maier2022surgical}
L.~Maier-Hein, M.~Eisenmann, D.~Sarikaya, K.~M{\"a}rz, T.~Collins, A.~Malpani, J.~Fallert, H.~Feussner, S.~Giannarou, P.~Mascagni \emph{et~al.}, ``Surgical data science--from concepts toward clinical translation,'' \emph{Medical image analysis}, vol.~76, p. 102306, 2022.

\bibitem{demir2023deep}
K.~C. Demir, H.~Schieber, T.~WeiseRoth, M.~May, A.~Maier, and S.~H. Yang, ``Deep learning in surgical workflow analysis: A review of phase and step recognition,'' \emph{IEEE Journal of Biomedical and Health Informatics}, 2023.

\bibitem{twinanda2016endonet}
A.~P. Twinanda, S.~Shehata, D.~Mutter, J.~Marescaux, M.~De~Mathelin, and N.~Padoy, ``Endonet: a deep architecture for recognition tasks on laparoscopic videos,'' \emph{IEEE transactions on medical imaging}, vol.~36, no.~1, pp. 86--97, 2016.

\bibitem{Liu_2021_CVPR}
D.~Liu, Q.~Li, T.~Jiang, Y.~Wang, R.~Miao, F.~Shan, and Z.~Li, ``Towards unified surgical skill assessment,'' in \emph{Proceedings of the IEEE/CVF Conference on Computer Vision and Pattern Recognition (CVPR)}, June 2021, pp. 9522--9531.

\bibitem{komatsu2024automatic}
M.~Komatsu, D.~Kitaguchi, M.~Yura, N.~Takeshita, M.~Yoshida, M.~Yamaguchi, H.~Kondo, T.~Kinoshita, and M.~Ito, ``Automatic surgical phase recognition-based skill assessment in laparoscopic distal gastrectomy using multicenter videos,'' \emph{Gastric Cancer}, vol.~27, no.~1, pp. 187--196, 2024.

\bibitem{ding2023less}
X.~Ding, X.~Yan, Z.~Wang, W.~Zhao, J.~Zhuang, X.~Xu, and X.~Li, ``Less is more: Surgical phase recognition from timestamp supervision,'' \emph{IEEE Transactions on Medical Imaging}, 2023.

\bibitem{zisimopoulos2018deepphase}
O.~Zisimopoulos, E.~Flouty, I.~Luengo, P.~Giataganas, J.~Nehme, A.~Chow, and D.~Stoyanov, ``Deepphase: surgical phase recognition in cataracts videos,'' in \emph{Medical Image Computing and Computer Assisted Intervention--MICCAI 2018: 21st International Conference, Granada, Spain, September 16-20, 2018, Proceedings, Part IV 11}.\hskip 1em plus 0.5em minus 0.4em\relax Springer, 2018, pp. 265--272.

\bibitem{kadkhodamohammadi2021towards}
A.~Kadkhodamohammadi, N.~Sivanesan~Uthraraj, P.~Giataganas, G.~Gras, K.~Kerr, I.~Luengo, S.~Oussedik, and D.~Stoyanov, ``Towards video-based surgical workflow understanding in open orthopaedic surgery,'' \emph{Computer Methods in Biomechanics and Biomedical Engineering: Imaging \& Visualization}, vol.~9, no.~3, pp. 286--293, 2021.

\bibitem{blum2010modeling}
T.~Blum, H.~Feu{\ss}ner, and N.~Navab, ``Modeling and segmentation of surgical workflow from laparoscopic video,'' in \emph{Medical Image Computing and Computer-Assisted Intervention--MICCAI 2010: 13th International Conference, Beijing, China, September 20-24, 2010, Proceedings, Part III 13}.\hskip 1em plus 0.5em minus 0.4em\relax Springer, 2010, pp. 400--407.

\bibitem{padoy2012statistical}
N.~Padoy, T.~Blum, S.-A. Ahmadi, H.~Feussner, M.-O. Berger, and N.~Navab, ``Statistical modeling and recognition of surgical workflow,'' \emph{Medical image analysis}, vol.~16, no.~3, pp. 632--641, 2012.

\bibitem{he2016deep}
K.~He, X.~Zhang, S.~Ren, and J.~Sun, ``Deep residual learning for image recognition,'' in \emph{Proceedings of the IEEE conference on computer vision and pattern recognition}, 2016, pp. 770--778.

\bibitem{szegedy2016rethinking}
C.~Szegedy, V.~Vanhoucke, S.~Ioffe, J.~Shlens, and Z.~Wojna, ``Rethinking the inception architecture for computer vision,'' in \emph{Proceedings of the IEEE conference on computer vision and pattern recognition}, 2016, pp. 2818--2826.

\bibitem{dosovitskiy2020image}
A.~Dosovitskiy, L.~Beyer, A.~Kolesnikov, D.~Weissenborn, X.~Zhai, T.~Unterthiner, M.~Dehghani, M.~Minderer, G.~Heigold, S.~Gelly \emph{et~al.}, ``An image is worth 16x16 words: Transformers for image recognition at scale,'' \emph{arXiv preprint arXiv:2010.11929}, 2020.

\bibitem{jin2017sv}
Y.~Jin, Q.~Dou, H.~Chen, L.~Yu, J.~Qin, C.-W. Fu, and P.-A. Heng, ``Sv-rcnet: workflow recognition from surgical videos using recurrent convolutional network,'' \emph{IEEE transactions on medical imaging}, vol.~37, no.~5, pp. 1114--1126, 2017.

\bibitem{czempiel2020tecno}
T.~Czempiel, M.~Paschali, M.~Keicher, W.~Simson, H.~Feussner, S.~T. Kim, and N.~Navab, ``Tecno: Surgical phase recognition with multi-stage temporal convolutional networks,'' in \emph{Medical Image Computing and Computer Assisted Intervention--MICCAI 2020: 23rd International Conference, Lima, Peru, October 4--8, 2020, Proceedings, Part III 23}.\hskip 1em plus 0.5em minus 0.4em\relax Springer, 2020, pp. 343--352.

\bibitem{czempiel2021opera}
T.~Czempiel, M.~Paschali, D.~Ostler, S.~T. Kim, B.~Busam, and N.~Navab, ``Opera: Attention-regularized transformers for surgical phase recognition,'' in \emph{Medical Image Computing and Computer Assisted Intervention--MICCAI 2021: 24th International Conference, Strasbourg, France, September 27--October 1, 2021, Proceedings, Part IV 24}.\hskip 1em plus 0.5em minus 0.4em\relax Springer, 2021, pp. 604--614.

\bibitem{ramesh2023dissecting}
S.~Ramesh, V.~Srivastav, D.~Alapatt, T.~Yu, A.~Murali, L.~Sestini, C.~I. Nwoye, I.~Hamoud, S.~Sharma, A.~Fleurentin \emph{et~al.}, ``Dissecting self-supervised learning methods for surgical computer vision,'' \emph{Medical Image Analysis}, vol.~88, p. 102844, 2023.

\bibitem{shi2021semi}
X.~Shi, Y.~Jin, Q.~Dou, and P.-A. Heng, ``Semi-supervised learning with progressive unlabeled data excavation for label-efficient surgical workflow recognition,'' \emph{Medical Image Analysis}, vol.~73, p. 102158, 2021.

\bibitem{yu2018learning}
T.~Yu, D.~Mutter, J.~Marescaux, and N.~Padoy, ``Learning from a tiny dataset of manual annotations: a teacher/student approach for surgical phase recognition,'' \emph{arXiv preprint arXiv:1812.00033}, 2018.

\bibitem{chen2018semi}
Y.~Chen, Q.~L. Sun, and K.~Zhong, ``Semi-supervised spatio-temporal cnn for recognition of surgical workflow,'' \emph{EURASIP Journal on Image and Video Processing}, vol. 2018, pp. 1--9, 2018.

\bibitem{yengera2018less}
G.~Yengera, D.~Mutter, J.~Marescaux, and N.~Padoy, ``Less is more: Surgical phase recognition with less annotations through self-supervised pre-training of cnn-lstm networks,'' \emph{arXiv preprint arXiv:1805.08569}, 2018.

\bibitem{shi2020lrtd}
X.~Shi, Y.~Jin, Q.~Dou, and P.-A. Heng, ``Lrtd: Long-range temporal dependency based active learning for surgical workflow recognition,'' \emph{International Journal of Computer Assisted Radiology and Surgery}, vol.~15, pp. 1573--1584, 2020.

\bibitem{bodenstedt2019active}
S.~Bodenstedt, D.~Rivoir, A.~Jenke, M.~Wagner, M.~Breucha, B.~M{\"u}ller-Stich, S.~T. Mees, J.~Weitz, and S.~Speidel, ``Active learning using deep bayesian networks for surgical workflow analysis,'' \emph{International journal of computer assisted radiology and surgery}, vol.~14, pp. 1079--1087, 2019.

\bibitem{zhang2022retrieval}
Y.~Zhang, S.~Bano, A.-S. Page, J.~Deprest, D.~Stoyanov, and F.~Vasconcelos, ``Retrieval of surgical phase transitions using reinforcement learning,'' in \emph{International conference on medical image computing and computer-assisted intervention}.\hskip 1em plus 0.5em minus 0.4em\relax Springer, 2022, pp. 497--506.

\bibitem{ward2021automated}
T.~M. Ward, D.~A. Hashimoto, Y.~Ban, D.~W. Rattner, H.~Inoue, K.~D. Lillemoe, D.~L. Rus, G.~Rosman, and O.~R. Meireles, ``Automated operative phase identification in peroral endoscopic myotomy,'' \emph{Surgical endoscopy}, vol.~35, pp. 4008--4015, 2021.

\bibitem{li2021temporal}
Z.~Li, Y.~Abu~Farha, and J.~Gall, ``Temporal action segmentation from timestamp supervision,'' in \emph{Proceedings of the IEEE/CVF Conference on Computer Vision and Pattern Recognition}, 2021, pp. 8365--8374.

\bibitem{behrmann2022unified}
N.~Behrmann, S.~A. Golestaneh, Z.~Kolter, J.~Gall, and M.~Noroozi, ``Unified fully and timestamp supervised temporal action segmentation via sequence to sequence translation,'' in \emph{European Conference on Computer Vision}.\hskip 1em plus 0.5em minus 0.4em\relax Springer, 2022, pp. 52--68.

\bibitem{khan2022timestamp}
H.~Khan, S.~Haresh, A.~Ahmed, S.~Siddiqui, A.~Konin, M.~Z. Zia, and Q.-H. Tran, ``Timestamp-supervised action segmentation with graph convolutional networks,'' in \emph{2022 IEEE/RSJ International Conference on Intelligent Robots and Systems (IROS)}.\hskip 1em plus 0.5em minus 0.4em\relax IEEE, 2022, pp. 10\,619--10\,626.

\bibitem{ijcai2023p77}
\BIBentryALTinterwordspacing
D.~Du, E.~Li, L.~Si, F.~Xu, and F.~Sun, ``Timestamp-supervised action segmentation from the perspective of clustering,'' in \emph{Proceedings of the Thirty-Second International Joint Conference on Artificial Intelligence, {IJCAI-23}}, E.~Elkind, Ed.\hskip 1em plus 0.5em minus 0.4em\relax International Joint Conferences on Artificial Intelligence Organization, 8 2023, pp. 690--698, main Track. [Online]. Available: \url{https://doi.org/10.24963/ijcai.2023/77}
\BIBentrySTDinterwordspacing

\bibitem{li2021anchor}
J.~Li and S.~Todorovic, ``Anchor-constrained viterbi for set-supervised action segmentation,'' in \emph{Proceedings of the IEEE/CVF Conference on Computer Vision and Pattern Recognition}, 2021, pp. 9806--9815.

\bibitem{huang2016connectionist}
D.-A. Huang, L.~Fei-Fei, and J.~C. Niebles, ``Connectionist temporal modeling for weakly supervised action labeling,'' in \emph{Computer Vision--ECCV 2016: 14th European Conference, Amsterdam, The Netherlands, October 11--14, 2016, Proceedings, Part IV 14}.\hskip 1em plus 0.5em minus 0.4em\relax Springer, 2016, pp. 137--153.

\bibitem{Souri_2022_BMVC}
\BIBentryALTinterwordspacing
Y.~Souri, Y.~A. Farha, E.~Bahrami, G.~Francesca, and J.~Gall, ``Robust action segmentation from timestamp supervision,'' in \emph{33rd British Machine Vision Conference 2022, {BMVC} 2022, London, UK, November 21-24, 2022}.\hskip 1em plus 0.5em minus 0.4em\relax {BMVA} Press, 2022. [Online]. Available: \url{https://bmvc2022.mpi-inf.mpg.de/0392.pdf}
\BIBentrySTDinterwordspacing

\bibitem{rahaman2022generalized}
R.~Rahaman, D.~Singhania, A.~Thiery, and A.~Yao, ``A generalized and robust framework for timestamp supervision in temporal action segmentation,'' in \emph{European Conference on Computer Vision}.\hskip 1em plus 0.5em minus 0.4em\relax Springer, 2022, pp. 279--296.

\bibitem{singhania2022iterative}
D.~Singhania, R.~Rahaman, and A.~Yao, ``Iterative contrast-classify for semi-supervised temporal action segmentation,'' in \emph{Proceedings of the AAAI Conference on Artificial Intelligence}, vol.~36, no.~2, 2022, pp. 2262--2270.

\bibitem{lin2017focal}
T.-Y. Lin, P.~Goyal, R.~Girshick, K.~He, and P.~Doll{\'a}r, ``Focal loss for dense object detection,'' in \emph{Proceedings of the IEEE international conference on computer vision}, 2017, pp. 2980--2988.

\bibitem{zhang2021surgical}
B.~Zhang, A.~Ghanem, A.~Simes, H.~Choi, and A.~Yoo, ``Surgical workflow recognition with 3dcnn for sleeve gastrectomy,'' \emph{International Journal of Computer Assisted Radiology and Surgery}, vol.~16, no.~11, pp. 2029--2036, 2021.

\bibitem{ramesh2021multi}
S.~Ramesh, D.~Dall’Alba, C.~Gonzalez, T.~Yu, P.~Mascagni, D.~Mutter, J.~Marescaux, P.~Fiorini, and N.~Padoy, ``Multi-task temporal convolutional networks for joint recognition of surgical phases and steps in gastric bypass procedures,'' \emph{International journal of computer assisted radiology and surgery}, vol.~16, pp. 1111--1119, 2021.

\bibitem{le2020transfer}
D.~N. Le, H.~X. Le, L.~T. Ngo, and H.~T. Ngo, ``Transfer learning with class-weighted and focal loss function for automatic skin cancer classification,'' \emph{arXiv preprint arXiv:2009.05977}, 2020.

\bibitem{qin2018weighted}
R.~Qin, K.~Qiao, L.~Wang, L.~Zeng, J.~Chen, and B.~Yan, ``Weighted focal loss: An effective loss function to overcome unbalance problem of chest x-ray14,'' in \emph{IOP Conference Series: Materials Science and Engineering}, vol. 428, no.~1.\hskip 1em plus 0.5em minus 0.4em\relax IOP Publishing, 2018, p. 012022.

\bibitem{gal2016dropout}
Y.~Gal and Z.~Ghahramani, ``Dropout as a bayesian approximation: Representing model uncertainty in deep learning,'' in \emph{international conference on machine learning}.\hskip 1em plus 0.5em minus 0.4em\relax PMLR, 2016, pp. 1050--1059.

\bibitem{guo2017calibration}
C.~Guo, G.~Pleiss, Y.~Sun, and K.~Q. Weinberger, ``On calibration of modern neural networks,'' in \emph{International conference on machine learning}.\hskip 1em plus 0.5em minus 0.4em\relax PMLR, 2017, pp. 1321--1330.

\bibitem{deng2009imagenet}
J.~Deng, W.~Dong, R.~Socher, L.-J. Li, K.~Li, and L.~Fei-Fei, ``Imagenet: A large-scale hierarchical image database,'' in \emph{2009 IEEE conference on computer vision and pattern recognition}.\hskip 1em plus 0.5em minus 0.4em\relax Ieee, 2009, pp. 248--255.

\bibitem{caron2021emerging}
M.~Caron, H.~Touvron, I.~Misra, H.~J{\'e}gou, J.~Mairal, P.~Bojanowski, and A.~Joulin, ``Emerging properties in self-supervised vision transformers,'' in \emph{Proceedings of the IEEE/CVF international conference on computer vision}, 2021, pp. 9650--9660.

\bibitem{lertnattee2004analysis}
V.~Lertnattee and T.~Theeramunkong, ``Analysis of inverse class frequency in centroid-based text classification,'' in \emph{IEEE International Symposium on Communications and Information Technology, 2004. ISCIT 2004.}, vol.~2.\hskip 1em plus 0.5em minus 0.4em\relax IEEE, 2004, pp. 1171--1176.

\bibitem{farha2019ms}
Y.~A. Farha and J.~Gall, ``Ms-tcn: Multi-stage temporal convolutional network for action segmentation,'' in \emph{Proceedings of the IEEE/CVF conference on computer vision and pattern recognition}, 2019, pp. 3575--3584.

\bibitem{graves2006connectionist}
A.~Graves, S.~Fern{\'a}ndez, F.~Gomez, and J.~Schmidhuber, ``Connectionist temporal classification: labelling unsegmented sequence data with recurrent neural networks,'' in \emph{Proceedings of the 23rd international conference on Machine learning}, 2006, pp. 369--376.

\bibitem{li2022recent}
J.~Li \emph{et~al.}, ``Recent advances in end-to-end automatic speech recognition,'' \emph{APSIPA Transactions on Signal and Information Processing}, vol.~11, no.~1, 2022.

\bibitem{alkendi2024advancements}
W.~AlKendi, F.~Gechter, L.~Heyberger, and C.~Guyeux, ``Advancements and challenges in handwritten text recognition: A comprehensive survey,'' \emph{Journal of Imaging}, vol.~10, no.~1, p.~18, 2024.

\bibitem{hannun2020differentiable}
A.~Hannun, V.~Pratap, J.~Kahn, and W.-N. Hsu, ``Differentiable weighted finite-state transducers,'' \emph{arXiv preprint arXiv:2010.01003}, 2020.

\bibitem{lavanchy2023challenges}
J.~L. Lavanchy, S.~Ramesh, D.~Dall’Alba, C.~Gonzalez, P.~Fiorini, B.~P. M{\"u}ller-Stich, P.~C. Nett, J.~Marescaux, D.~Mutter, and N.~Padoy, ``Challenges in multi-centric generalization: phase and step recognition in roux-en-y gastric bypass surgery,'' \emph{International journal of computer assisted radiology and surgery}, pp. 1--9, 2024.

\end{thebibliography}


\end{document}